\documentclass[journal]{IEEEtran}

\ifCLASSINFOpdf
\else
   \usepackage[dvips]{graphicx}
\fi
\usepackage{url}

\usepackage{graphicx}
\usepackage{bm}
\usepackage{amsmath}
\usepackage{microtype}
 \usepackage{multirow}
\usepackage{float}
\usepackage{overpic}
\usepackage{booktabs}
\usepackage{hyperref}
\hypersetup{hypertex=true,
	colorlinks=true,
	linkcolor=blue,
	anchorcolor=blue,
	citecolor=blue,
urlcolor=blue}
\begin{document}

\title{An Improved Normed-Deformable Convolution \\ for Crowd Counting}

\author{Xin Zhong, Zhaoyi Yan, Jing Qin, Wangmeng Zuo and Weigang Lu\vspace{-1em} 

\thanks{Xin Zhong is with the Department of Educational Technology, Ocean University of China, Qingdao, 266000, China. (zhongxin@stu.ouc.edu.cn). Zhaoyi Yan is with the School of Computer Science and Technology, Harbin Institute of Technology, Harbin, 150001, China. (yanzhaoyi@outlook.com). Jing Qin is with the Department of Educational Technology, Ocean University of China, Qingdao, 266000, China. (qinjing@stu.ouc.edu.cn). Wangmeng Zuo is with the School of Computer Science and Technology, Harbin Institute of Technology, Harbin, 150001, China. (cswmzuo@hit.edu.cn). Weigang Lu is with the Department of Educational Technology, Ocean University of China, Qingdao, 266000, China. (luweigang@ouc.edu.cn). *Correspondence should be addressed to Weigang Lu; luweigang@ouc.edu.cn}}

\markboth{}
{Shell \MakeLowercase{\textit{et al.}}: Bare Demo of IEEEtran.cls for IEEE Journals}
\maketitle

\begin{abstract}
In recent years, crowd counting has become an important issue in computer vision. In most methods, the density maps are generated by convolving with a Gaussian kernel from the ground-truth dot maps which are marked around the center of human heads. Due to the fixed geometric structures in CNNs and indistinct head-scale information, the head features are obtained incompletely. Deformable convolution is proposed to exploit the scale-adaptive capabilities for CNN features in the heads. By learning the coordinate offsets of the sampling points, it is tractable to improve the ability to adjust the receptive field. However, the heads are not uniformly covered by the sampling points in the deformable convolution, resulting in loss of head information. To handle the non-uniformed sampling, an improved Normed-Deformable Convolution (\textit{i.e.,}NDConv) implemented by Normed-Deformable loss (\textit{i.e.,}NDloss) is proposed in this paper. The offsets of the sampling points which are constrained by NDloss tend to be more even. Then, the features in the heads are obtained more completely, leading to better performance. Especially, the proposed NDConv is a light-weight module which shares similar computation burden with Deformable Convolution. In the extensive experiments, our method outperforms state-of-the-art methods on ShanghaiTech A, ShanghaiTech B, UCF\_QNRF, and UCF\_CC\_50 dataset, achieving 61.4, 7.8, 91.2, and 167.2 MAE, respectively. The code is available at \url{https://github.com/bingshuangzhuzi/NDConv}.
\end{abstract}


\IEEEpeerreviewmaketitle

\vspace{-0.2cm}
\section{Introduction}

\IEEEPARstart{D}{ue} to the expansion of worldwide population and the acceleration of urbanization, the probability of crowd massing has increased in recent years. In some scenarios, stampedes and large-scale events pose a hazard to public safety that should be prevented all the time. In order to keep the public safety, crowd counting gains considerable attention in the field of computer vision.
Furthermore,  the foundation of hardware for crowd counting is provided by the advancement of remote-distance intellectualized cameras \cite{6,7,8},\cite{40,42,41,43}. Therefore, there are many researches based on crowd counting\cite{1,2,3} which estimate the amount of crowd in campuses, shopping malls, and train stations accurately. 

Currently, the previous crowd counting algorithms can be classified into three categories: tracking-based 
\begin{figure}[h]
	\centerline{\includegraphics[width=\columnwidth]{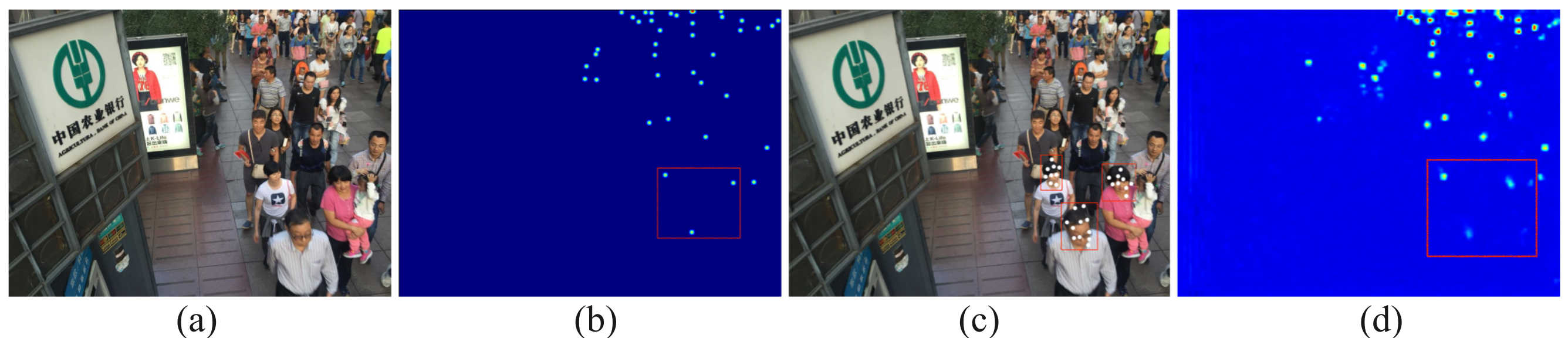}}
	\setlength{\abovecaptionskip}{-0.1cm}
	\caption{Visualization of the offsets and density maps predicted from models trained with DConv. (a) is one of the input images in ShanghaiTech B dataset, (b) is the ground\_truth, the offsets are visualized in (c), and the estimated density map is shown in (d). Due to the sampling points are uniformed in (c), (d) shows less regular Gaussian blobs.}
	\label{fig:fig1}
	\vspace{-0.6cm}
\end{figure}
methods\cite{9,10}, feature-based regression methods\cite{11,12}, and CNN-based methods\cite{13,14,15}.  By comparison, CNN-based methods show their superiority in higher accuracy, efficiency, and robustness. Although great progress has been made in prior works of crowd counting implemented by convolutional neural networks (CNNs), the inability of CNNs adapting to alterations of head-scale\cite{37,38,39} is still a barrier, restricting the enhancement of counting accuracy. The methods\cite{16,17} based on multi-scale feature fusion attempt to address above-mentioned problems. However, there are also some obstacles. Firstly, due to the limitation that parameters of network and computational time balloon along with the increase of the scales of features, the training time is unbearable. In addition, only a few works pay attention on the mechanisms to handle geometric transformations.

To address the above-mentioned problems, a module called Deformable Convolution(\textit{i.e.,}DConv)\cite{18,19} is proposed to enhance the CNNs' capability of modeling geometric transformations by adding additional offsets. However, as shown in the  Fig.~\ref{fig:fig1}~, it is still non-trivial to sample the head features correctly and uniformly due to the uncontrollable offsets of sampling points in the above-mentioned versions. Based on the preceding discussion, we argue that it is difficult to sample uniformly by simply integrating Deformable Convolution into the network.  Therefore, in this work, we attempt to address the problem of non-uniform sampling with extra loss.

Since the offsets are not constrained in DConv, the sampling region is irregular, making it difficult to sample the features of heads uniformly. Thus, a novel Normed-Deformable Convolution (\textit{i.e.,}NDConv) which is implemented by Normed-Deformable loss (\textit{i.e.,}NDloss) is proposed to address the forementioned problem. The offsets of sampling points are controlled by NDloss. As a result, the features in the heads are captured more completely, leading to better performance. In our implementation, CSRNet\cite{20} is adopted as the backbone. As the baseline, the last dilated convolution layer is replaced with Deformable Convolution which is denoted as CSRNet$\star$. Without bells and whistles, by replacing the Deformable Convolution in CSRNet$\star$ with the proposed NDConv, we achieve better performance (\textit{i.e.,}
$5.5\%$, $13.3\%$, $4.5\%$ and $4.2\%$  improvement in MAE(Mean Absolute Error) on ShanghaiTech A, ShanghaiTech B, UCF\_QRNF, and UCF\_CC\_50.)

In summary, the main contributions of this paper are three-fold. 
\begin{itemize}
	\item [1)]
	Comparing with conventional Deformable Convolution, a novel Normed-Deformable Convolution (\textit{i.e.,}NDConv) is proposed in which the sampling points tend to be more uniformly spread over the head, resulting in more adequate feature aggregation.
	
	\item [2)]
	Our NDconv encourages more thoughts that shape priors should be intergrated into the Deformable Convolution by constraining the sampling offsets.
	
	\item [3)]
	Based on the proposed NDConv, better performance is achieved on several crowd counting datasets than the state-of-the-art methods.
\end{itemize}

\vspace{-0.4cm}
\section{Proposed Method}

In this section, the definition and principles of Normed-Deformable Loss are described firstly, and then the architecture of the training network is presented.
\vspace{-0.5cm}
\subsection{Normed-Deformable Loss}
Given a convolution of kernel size \textls[-400]{$3\times3$} for example, the standard nine sampling points respectively denoted as $a,b,c,d,e,f,g,h,i$ lie strictly on the grid.
For Deformable Convolution, due to the existence of sampling offsets, the sampling points $a,b,c,d,e,f,g,h,i$ move to $A,B,C,D,E,F,G,H,I$, respectively.
In this case, the offsets are two-dimensional and are recorded as $(\Delta A_x,\Delta A_y), \cdots, (\Delta I_x, \Delta I_y)$, where we have
$\mathrm{\it A}=\Delta {\it A}+{\it a},\cdots,\mathrm{\it I}=\Delta {\it I}+{\it i}$.

Fig.~\ref{fig:fig2}(a) visualizes a typical case of sampling points in DConv, the sampling points travel almost randomly due to no explicit constraints applied on the corresponding offsets.
%
%
Obviously, the sampling positions fail to evenly cover the head region, leading to inadequate feature aggregation.
By considering the prior that the head tends to be in the shape of ellipse, an intuitive thought is constraining the offsets of sampling points to be in accordance with the prior, making the feature aggregation more sufficient.

Here, we make a simplification on the prior.
Specifically, the ellipse is decoupled into four parallelograms, \textit{i.e.,}$ABED$, $CBEF$, $GHED$, $IHEF$, depicted in Fig.~\ref{fig:fig2}(b).
Before we build the corresponding loss for these parallelograms, three restrictions should be constructed to make the whole sampling points more reasonable.
(i) The central sampling point $E$ should be close to $e$, which is intuitive.
(ii) $D$ and $F$ share the same distance away from $e$, and both $D$ and $F$ should be close to $x$-axis.
(iii) Similarly, $B$ and $H$ should be the same distance away from $e$, and both $B$ and $H$ should be close to $y$-axis.

For restriction (i), we have:
\begin{equation}
	\mathcal{L}_{e}=\left\|\Delta E_{x}\right\|_{2}^{2}+\left\|\Delta E_{y}\right\|_{2}^{2} \\
\end{equation}

For restriction (ii), we have:
\begin{equation}
	\mathcal{L}_{h o r}=\left\|\Delta D_{x}+\Delta F_{x}\right\|_{2}^{2}+\left\|\Delta D_{y}\right\|_{2}^{2}+\left\|\Delta F_{y}\right\|_{2}^{2}
\end{equation}

And for restriction (iii), we have:
\begin{equation}
	\mathcal{L}_{v e c}=\left\|\Delta B_{y}+\Delta H_{y}\right\|_{2}^{2}+\left\|\Delta B_{x}\right\|_{2}^{2}+\left\|\Delta H_{x}\right\|_{2}^{2}
\end{equation}

Finally, we constrian the four diagnoal points $A$, $C$, $G$, $I$, making them close to the four parallelograms, \textit{i.e.,}$ABED$, $CBEF$, $GHED$, $IHEF$:
\vspace{-0.5cm}
\begin{figure}[h]
	\centerline{\includegraphics[width=\columnwidth]{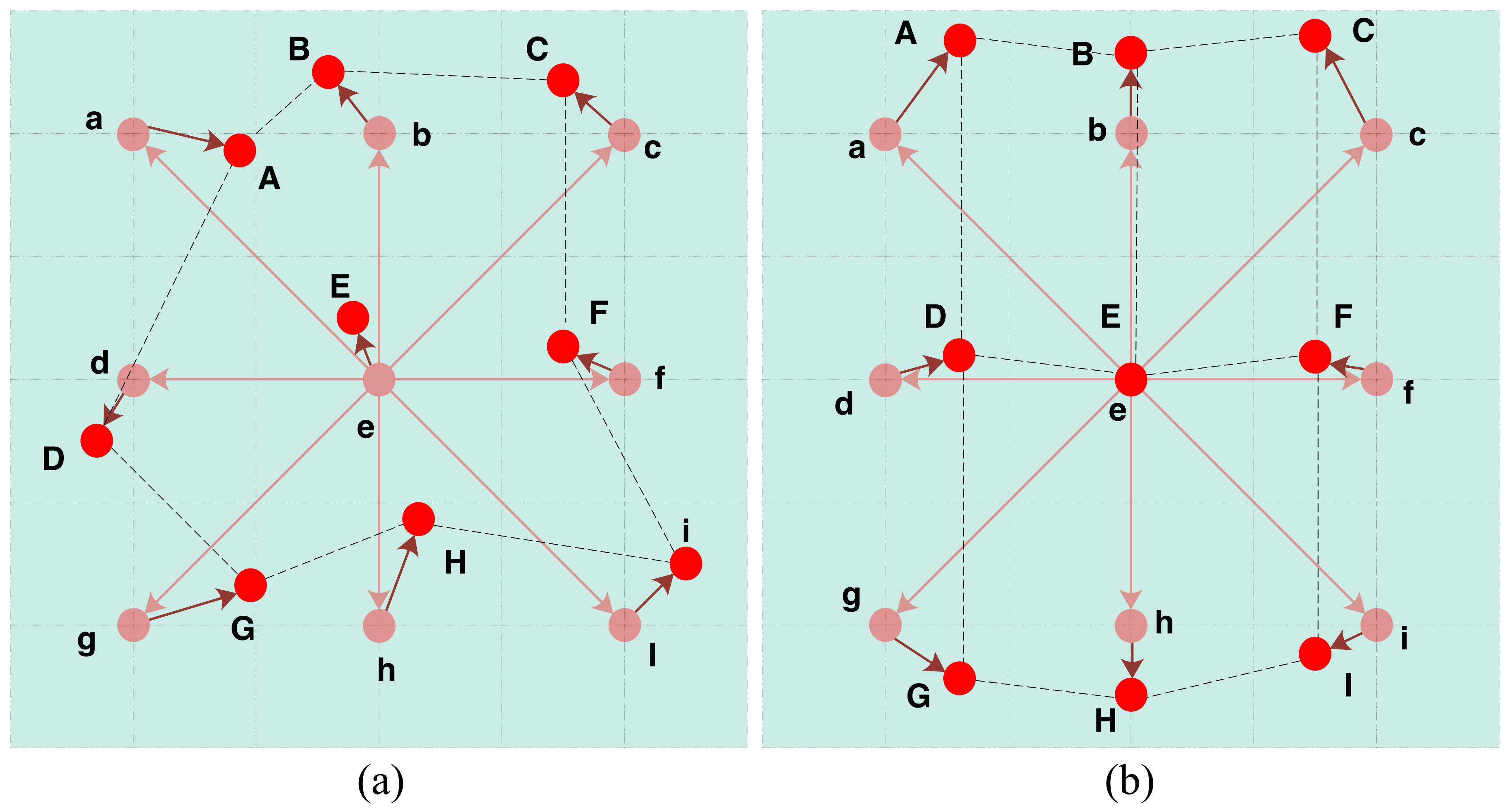}}
	\setlength{\abovecaptionskip}{-0.1cm}
	\caption{Illustrations of conventional DConv and our NDConv, as demonstrated in (a) and (b). The sampling points are represented by the balls. Among them, the pink balls represent the sampling points of the conventional convolution (indicated by lowercase letters, {\it a-i}), the red balls represent the actual sampling points (indicated by uppercase letters {\it A-I}), and the offsets are represented by the dark purple arrows.}
	\label{fig:fig2}
	\vspace{-0.4cm}
\end{figure}
\begin{equation}
	\begin{aligned}
		\mathcal{L}_{A} =\|(d+\Delta D)+(b+\Delta B)-(e+\Delta E)-a\| _{2}^{2}\\
		\mathcal{L}_{C} =\|(f+\Delta F)+(b+\Delta B)-(e+\Delta E)-c\| _{2}^{2}\\
		\mathcal{L}_{G} =\|(d+\Delta D)+(h+\Delta H)-(e+\Delta E)-g\| _{2}^{2}\\
		\mathcal{L}_{I} =\|(f+\Delta F)+(h+\Delta H)-(e+\Delta E)-i\|_{2}^{2}
	\end{aligned}
\end{equation}

Thus, NDloss can be presented as Eq.\eqref{eqn:nd_loss}:
\begin{equation}\label{eqn:nd_loss}
	\mathcal{L}_{\text {nd}}=\mathcal {L}_{e}+\mathcal {L}_{\text {hor }}+\mathcal {L}_{\text {vec }}+\mathcal {L}_{A}+\mathcal {L}_{C}+\mathcal {L}_{G}+\mathcal {L}_{I}
\end{equation}

Then, the loss of density estimation is:
\begin{equation}
	\mathcal{L}_{d e n}=\min _{\Phi} \frac{1}{2 N}\left\|P\left(I_{i} ; \boldsymbol{\Phi}\right)-Y_{i}\right\|_{2}^{2}
\end{equation}
{\it Y$_i$ }is the density map of ground\_truth, $\boldsymbol{\Phi}$ denotes the parameters of the network, $P\left(I_{i} ; \boldsymbol{\Phi}\right)$
represents the estimation of density map, and $N$ is the batch size.
The final loss is:
\begin{equation}
	\mathcal {L}_{a l l}=\mathcal {L}_{d e n}+\lambda \mathcal {L}_{nd}
\end{equation}
where $\lambda$ is the super-parameter to balance the two losses.

\subsection{Net Architecture}
\begin{figure*}[htbp] 
	\centering
	\includegraphics[width=\linewidth,height=6cm]{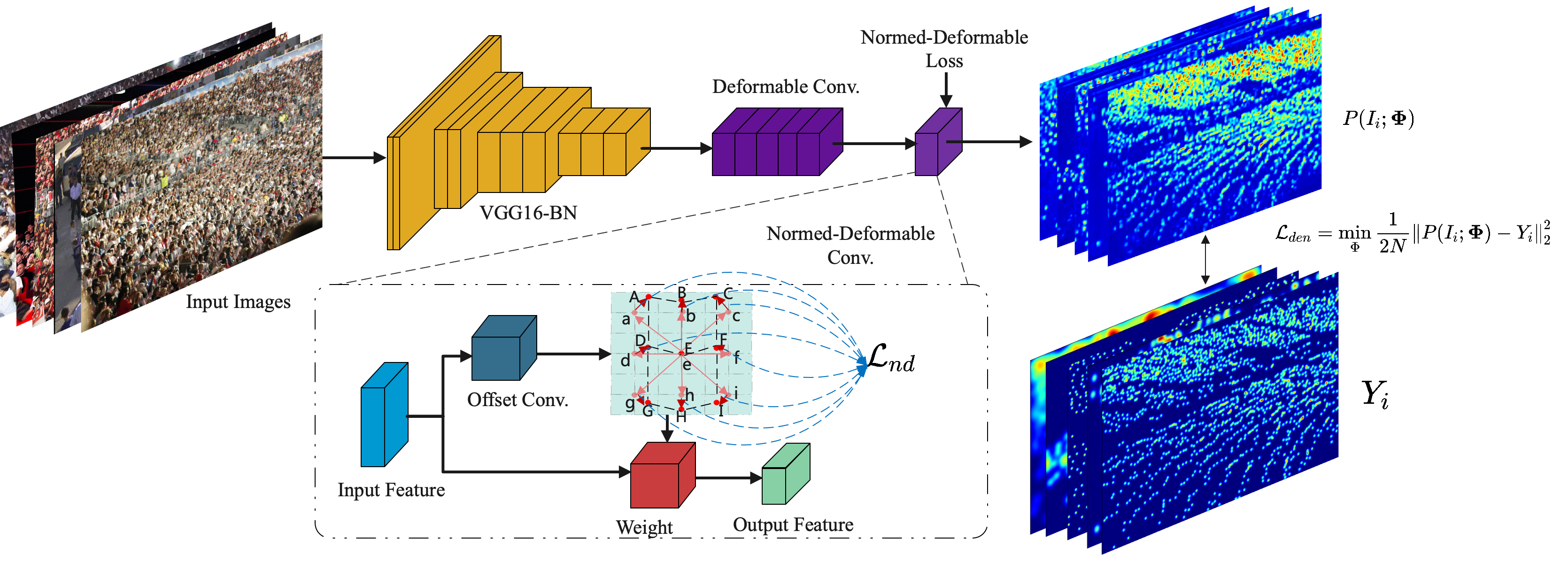}
	\setlength{\abovecaptionskip}{-0.4cm}
	\caption{\it The architecture of the proposed NDConv network. We adopt CSRNet$\star$ as our backbone, where the last layer of dilated convolution is replaced with NDConv to constrain the sampling points.}
	\label{fig:fig3}
	\vspace{-0.5cm}
\end{figure*}

The architecture of the network is illustrated in Fig.~\ref{fig:fig3}.
CSRNet\cite{20} is adopted as the backbone network, specifically, we add Batch Normalziation 
layer after each convolution layer to enhance the robustness of training with cropped images.
After that, the last layer of dilated convolution is replaced with DConv or NDConv, denoted as CSRNet$\star$ and CSRNet (NDConv), respectively. In the next section, the effectiveness of our NDConv is verified. 

\section{Experimental results}
In this section, four datasets are adopted in our experiments, including ShanghaiTech A, ShanghaiTech B\cite{5}, UCF-QRNF\cite{21}, and UCF\_CC\_50\cite{22}. Firstly, the evaluation metric and datasets are introduced, and then the implementation details are described. Subsequently, the proposed NDconv is compared with state-of-the-art methods on the four above-mentioned datasets. Finally, the ablation study is conducted to explain the meaning of parameters in experiments.

\subsection{Datasets and Implementation Details}

\subsubsection{Datasets}
On four popular datasets, ShanghaiTech Part A, ShanghaiTech Part B, UCF-QNRF, and UCF\_CC\_50, the proposed NDConv and other state-of-the-art approaches are evaluated.

\noindent{\bfseries ShanghaiTech}\cite{5}. It is composed of part A and part B. For Part A, the images are obtained from the web, and the images in Part B are views of Shanghai streets. Part A contains $482$ crowd images. Different from Part A, there are more high-resolution images in Part B. It contains $716$ images, while the counts of crowd in per image is from $9$ to $578$.

\noindent{\bfseries UCF-QNRF}\cite{21}. Compared with ShanghaiTech, this is a larger crowd counting dataset that contains $1, 535$ high-resolution images and $1.25$ million head annotations. In this dataset, there are $1,201$ images of extremely congested scenes for training and the rest are test images. The maximum count of an image is $12, 865$. 

\noindent{\bfseries UCF\_CC\_50}\cite{22}. In our experiment, there is also a challenging dataset called UCF\_CC\_50 which contains $50$ images. Although the number of images is small, the count in per image varies drastically from $94$ to $4,543$. Following the setting in \cite{22}, we divide the dataset into five subsets, and then $5$-fold cross-validation is performed.

\subsubsection{Evaluation Metric}
{Mean Absolute Error (MAE) and Mean Square Error (MSE) are adopted as the evaluation metrics, same as the previous works \cite{23,24,25}.

\subsubsection{Implementation Details}
To improve the robustness of training with cropped images, the backbone of CSRNet\cite{20} is replaced with VGG16-BN and we insert Batch-Normalization\cite{44} layer after each dilated convolution. Then, the last layer of dilated convolution is replaced with DConv. {\bfseries After these changes, the model is termed as CSRNet$\star$ and is the baseline of experiments. Subsequently, the last layer of DConv is replaced with NDconv, denoted as Ours (NDConv).}

In the experiments, we use a fixed $\! 7 \times \! 7$ Gaussian kernel to generate density maps and Adam\cite{27} optimizer is used with a learning rate of $1${\it e}-$4$. All images are resized to $\! 400 \times \! 400$. The batch size and parameter $\lambda$ are set to $4$ and $0.001$. We train $400$ epochs in two days. 

\subsection{Evaluations and Comparisons}
The results of experiments on ShanghaiTech A and B are shown in Table~\ref{table:table1}. 
\vspace{-0.2cm}
\begin{table}[h]
	\caption{Comparisons on Shanghaitech A and B}
	\centering
	\renewcommand\arraystretch{1}  
	\label{table:table1}
	\small
	\setlength{\tabcolsep}{2pt}
	\begin{tabular}{l|l|cc|cc}
		\toprule[1pt]
		\multicolumn{1}{c|}{\multirow{2}{*}{Methods}} & \multirow{2}{*}{Venue\&Year} & \multicolumn{2}{l|}{ShanghaiTechA} & \multicolumn{2}{l}{ShanghaiTechB} \\ \cline{3-6} 
		\multicolumn{1}{c}{} &                        & MAE  & MSE   & MAE  & MSE  \\ \hline
		CSRNet{\cite{20}}          & CVPR2018               & $68.2$ & $115.0$ & $10.6 $& $16.0$ \\
		SANet{\cite{30}}            & CVPR2018               &$ 67.0$ & $104.5$ &$ 8.4 $ & $13.6$ \\
		TEDnet{\cite{31}}           & CVPR2019               & $64.2$ & $109.1 $&$ 8.2 $ &$ 12.8$ \\
		ADCrowdNet{\cite{25}}      & CVPR2019               & $63.2$ &$ 98.9 $ & $8.2 $ &$ 15.7$ \\
		PSDDN{\cite{32}}            & CVPR2019               &$ 65.9$ &$ 112.3$ &$ 9.1$  &$ 14.2 $\\
		HA-CNN{\cite{28}}               & TIP2019               & $62.9 $& \textbf{94.9 }& $8.1$  &$ 13.4$ \\
		LSC-CNN{\cite{29}}              & TPAMI2020               & $66.4$ &$ 117.0 $ &$8.1 $ &\textbf{12.7}\\
		DensityCNN{\cite{33}}        & TMM2020                &$ 63.1$ &$ 106.3$ &$ 9.1 $ &$ 16.3 $\\
		DENet{\cite{34}}            & TMM2020                & $65.5 $& $101.2$ & $9.6$  & $15.4$ \\ \hline
		CSRNet$\star$(Baseline)    & \multicolumn{1}{c|}{-} & $65.0$ &$102.63$       & $9.0$  &$15.3  $    \\
		Ours(NDConv)          & \multicolumn{1}{c|}{-} & \textbf{61.4} &$104.18 $      &\textbf{7.8}  &$13.8$   \\ \bottomrule[1pt]     
	\end{tabular}
\vspace{-0.3cm}
\end{table}

Compared with state-of-the-art, our method performs the best, indicating the advantages of constrained offsets in deform convolution. For ShanghaiTech A, our method achieves the best $61.4$ MAE with $2.5$ performance improvement against the baseline CSRNet$\star$. The result of evaluation on ShanghaiTech B has also lived up to expectations, our method achieves $7.8$ MAE and $13.3$ relative MAE decrease with the baseline.
\vspace{-0.3cm}
\begin{table}[h]
	\caption{Comparisons on UCF-QNRF and UCF\_CC\_50}
	\centering
	\renewcommand\arraystretch{1}  
	\label{table:table2}
	\small
	\setlength{\tabcolsep}{3pt}
	\begin{tabular}{l|l|cc|cc}
		\toprule[1pt]
		\multicolumn{1}{c|}{\multirow{2}{*}{Methods}} & \multirow{2}{*}{Venue\&Year} & \multicolumn{2}{l|}{UCF-QNRF} & \multicolumn{2}{l}{UCF\_CC\_50} \\ \cline{3-6} 
		\multicolumn{1}{c}{} &                        & MAE   & MSE   & MAE   & MSE   \\ \hline
		CSRNet{\cite{20}}           & CVPR2018               & -     & -     &$ 266.1$ & $397.5$  \\
		SANet{\cite{30}}              & CVPR2018               & -     & -     & $258.4$ & $334.9$  \\
		TEDnet{\cite{31}}           & CVPR2019               & $113 $  &$ 188  $ & $294.4$ & $354.5 $ \\
		ADCrowdNet{\cite{25}}      & CVPR2019               & -     & -     &$ 257.1$ & $363.5$ \\
		PSDDN{\cite{32}}           & CVPR2019               & -     & -     & $359.4$ & $514.8$ \\
		HA-CCN{\cite{28}}              & TIP2019              & $118.1$  & $180.4 $& $256.2 $    & $348.4  $   \\
		LSC-CNN{\cite{29}}             & TPAMI2020               & $120.5$ & $218.2$ & $225.6$ & $302.7 $\\
		DensityCNN{\cite{33}}      & TMM2020                & $101.5$ & $186.9$ & $244.6$ & $341.8$ \\
		DENet{\cite{34}}            & TMM2020                & -     & -     & $241.9$ & $345.4$ \\
		DS-CNN{\cite{50}}            & AJSE2021               & $115.2$     & $175.7$     & $229.4$ & $325.6$ \\
		SDIH-DDM{\cite{51}}            & TVCJ2020               & $112$     & $173$     & - & - \\
		\hline
		CSRNet$\star$(Baseline)    & \multicolumn{1}{c|}{-} & $95.5$  & \textbf{165.3}  & $174.6 $  & \textbf{237.0}  \\
		Ours(NDConv)          & \multicolumn{1}{c|}{-} & \textbf{91.2}  & $165.6$  & \textbf{167.2}   & $240.6$ \\ \bottomrule[1pt]
	\end{tabular}
\end{table}

For the datasets, UCF-QNRF and UCF\_ CC\_50, the comparison between our methods and state-of-the-arts is recorded in Table~\ref{table:table2}. Compared with the baseline CSRNet$\star$, our method achieves significant gain on MAE. Respectively, $91.2$ and $167.2$ MAE are achieved on datasets UCF-QNRF and UCF\_CC\_50, with $4.5$\%, $4.2$\% performance improving. In particular, on the premise of great performance improvement on CSRNet$\star$, our method further reduces MAE, indicating the effectiveness of the constrained offsets in the dense scene.
\vspace{-0.4cm}
\subsection{Ablation Study}\label{formats}
In this section, the influence of the number of deformable layers in our network is demonstrated firstly. On the other hand, visualization of the constrained offsets is shown in  Fig.~\ref{fig:fig4}~ and the validity of NDloss is verified. Finally, the extensibility of NDConv on another backbone is indicated.
\subsubsection{Influence of the number of deformable layers}  There are $6$ layers of dilated convolution integrated into the architecture of CSRNet. Each layer of dilated convolution is changed into Deformable Convolution gradually. Then, each layer of Deformable Convolution is replaced with NDConv. The performance is shown in Table~\ref{table:table3}. While the number of deformable layers is increased, the performance of the above-mentioned experiment is decreased. The peak values of baseline are from $172.9$ to $187.4$ MAE, while the trend in performance of NDConv is similar, which is from $167.2$ to $184.6$ MAE on UCF\_CC\_50. Thus, the effectiveness is better by only replacing the last layer of dilated convolution with DConv and NDConv respectively in the network.
\vspace{-0.3cm}
\begin{table}[h]
	
	\caption{Effects of the number of deformable layers on UCF\_CC\_50}
	\label{table:table3}
	\renewcommand\arraystretch{1}  
	\small
	\centering
	\setlength{\tabcolsep}{3pt}
	\begin{tabular}{c|cc|cc}
		\toprule[1pt]
		\multirow{2}{*}{Deformable Layers} & \multicolumn{2}{c|}{CSRNet$\star$} & \multicolumn{2}{c}{Ours(NDConv)} \\ \cline{2-5} 
		& MAE & MSE & MAE & MSE \\ \hline
		1 & $174.6 $   &$237.0 $   &\textbf{167.2}     &$240.6  $   \\
		2 &\textbf{172.9}     &$268.3$     &$185.0  $  &$296.1  $   \\
		3 & $175.4$    &\textbf{233.3}     &$168.2  $   &\textbf{235.4}     \\
		4 &$182.7  $   &$271.2   $  &$177.2  $   &$245.9$     \\
		5 &$187.4 $    &$266.5 $    &$177.6 $    &$235.7 $    \\
		6 & $174.9 $   &$256.2$     &$184.6 $    &$263.3 $    \\ \bottomrule[1pt]
	\end{tabular}
\end{table}
\vspace{-0.3cm}
\subsubsection{Visualization of offsets}
Experiments have shown that our NDConv surpasses Deformable Convolution and the visualization of the offsets in Fig.~\ref{fig:fig4} as previous work\cite{36} also validates our hypothesis. %
Fig.~\ref{fig:fig4}(c) and Fig.~\ref{fig:fig4}(f) demonstrate the sampling points of DConv and our proposed NDConv.
It is obvious that the sampling points of NDConv tend to be spread more uniformly over the heads, leading to more sufficient feature aggregation.
Accordingly, Fig.~\ref{fig:fig4}(e) shows more regular Gaussian blobs than Fig.~\ref{fig:fig4}(d), which further verifies the reasonableness of our NDConv.

\begin{figure}[h]
	\centerline{\includegraphics[width=\columnwidth]{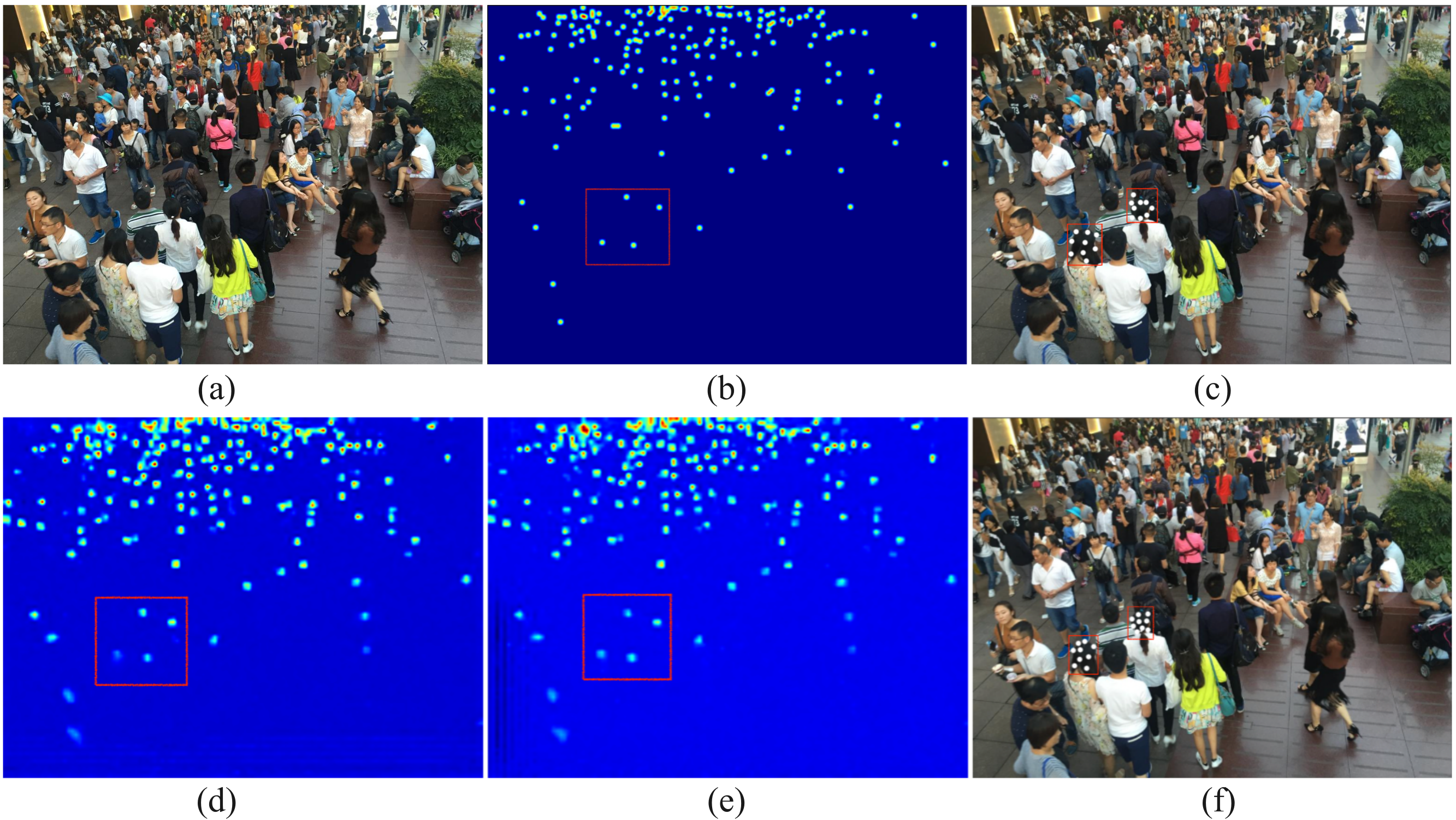}}
	\setlength{\abovecaptionskip}{-0.2cm}
	\caption{Visualization of an example from ShanghaiTech B dataset. (a) is the input image; (b) is the ground\_truth; (c) and (f) are the visualization of offsets in DConv and NDConv; the predicted density map in DConv and NDConv are shown in (d) and (e).}
	\label{fig:fig4}
	\vspace{-0.5cm}
\end{figure}

\subsubsection{Extensibility to Another Backbone} A new baseline consisted of ResNet-50\cite{35} is built to verify the extensibility of NDConv. Firstly, ResNet-50 (backbone) with the four dilated convolutions is not changed. Then, the last layer of dilated convolutions is replaced with Deformable Convolution as the new baseline called ResNet-50 (DConv). Finally, the layer of Deformable Convolution is replaced with NDConv in ResNet-50 (NDConv). The comparisons of results on UCF\_QNRF and UCF\_CC\_50 are shown in Table~\ref{table:table4}. For ResNet-50 (backbone), we get  $99.3$ and $227.6$ MAE on UCF\_QNRF and UCF\_CC\_50,  respectively. When the last layer of convolution is replaced with Deformable Convolution, the baseline delivers $98.0$ and $231.3$ MAE. Particularly, if the layer of Deformable Convolution is replaced by NDConv, a better performance is achieved $95.4$ and $222.4$ MAE. The extensibility and effectiveness of NDConv on another backbone are indicated by the results of the experiment.
\vspace{-0.3cm}
\begin{table}[h]
	\setlength{\abovecaptionskip}{-0.1cm}
	\caption{Comparisons on the ResNet-50 backbone}
	\label{table:table4}
	\small
	\centering
	\setlength{\tabcolsep}{3pt}
	\begin{tabular}{c|cc|cc}
		\toprule[1pt]
		\renewcommand\arraystretch{1}   
		\multirow{2}{*}{Methods} &
		\multicolumn{2}{c}{\begin{tabular}[c]{@{}c@{}}UCF\_QNRF\\[0.5ex]\end{tabular}} &
		\multicolumn{2}{c}{\begin{tabular}[c]{@{}c@{}}UCF\_CC\_50\\[0.5ex]\end{tabular}} \\ \cline{2-5} 
		& MAE  & MSE   & MAE  & MSE  \\[0.5ex] \hline
		ResNet-50(backbone) &$ 99.3$ & $181.5$ & $227.6$&\textbf{307.6 } \\
		ResNet-50(DConv)    &$98.0 $  &\textbf{178.0 }    &$ 231.3 $   & $317.0 $  \\
		ResNet-50(NDConv)   &\textbf{95.9}   &$182.4   $  & \textbf{222.4}    & $315.4 $  	 \\ \bottomrule[1pt]

	\end{tabular}
\end{table}

\subsubsection{The effects of $\mathcal{L}_{nd}$} The final loss consists of $\mathcal{L}_{nd}$ and $\mathcal{L}_{d e n}$.  In this section, the effect of super-parameter $\lambda$ is investigated. The performance varies with the values of $\mathcal{L}_{nd}$. By conducting the ablation study, it is verified that when the weight of the super-parameter $\lambda$ is $1${\it e}-$3$, the better performance is achieved. The experimental data is shown in Table~\ref{table:table5}. 
\vspace{-0.2cm}
\begin{table}[h]
	\setlength{\abovecaptionskip}{-0.1cm}
	\caption{Effects of the weight of $\mathcal{L}_{nd}$ on UCF-QNRF and UCF\_CC\_50}
	\label{table:table5}
	\small
	\centering
	\setlength{\tabcolsep}{3pt}
	\begin{tabular}{c|c|ccccc}
		\toprule[1pt]
		\multirow{2}{*}{Datasets}            & \multirow{2}{*}{Metrics} & \multicolumn{5}{c}{$\lambda$(super-parameter)} \\ \cline{3-7} 
		&                          & $1${\it e}-$1$    & $1${\it e}-$2$  & $1${\it e}-$3$   & $1${\it e}-$4$  & $0$  \\ \hline
		\multirow{2}{*}{UCF-QNRF\cite{21}}    & MAE                      & $94.8$    &$95.6$      &\textbf{91.2}  & $93.8$     & $95.5$  \\
		& MSE                      & $167.4$   & \textbf{165.2}    &$165.6$  & $167.2$     & $165.3$ \\
		\multirow{2}{*}{UCF\_CC\_50\cite{22}} &MAE                      & $189.5$      & $182.4$    &\textbf{167.2}  & $175.2$     & $174.6$  \\
		& MSE                      & $264.2$       & $285.1$   & $240.6$  & $238.7$    & \textbf{237.0}  \\ \bottomrule[1pt]
	\end{tabular}
\end{table}
\vspace{-0.3cm}
\section{Conclusion}

Normed-Deformable Convolution(NDConv) is proposed in this paper, which shows its specificity in the constrained sampling offsets via NDloss.
NDloss is designed under the guidance of the shape of head, so as to enhance the effectiveness of feature aggregation.
Experiments show that our proposed NDConv surpasses the conventional Deformable Convolution over four popular datasets with little extra computational burden.
\vspace{-0.3cm}
\section*{Acknowledgements}
This work has been supported by the Natural Science Foundation of Shandong Province (No. ZR2021MF011).

\end{document}